\documentclass{article}

    \PassOptionsToPackage{numbers, compress}{natbib}

 \usepackage[preprint]{neurips_2022}

\usepackage[utf8]{inputenc} 
\usepackage[T1]{fontenc}    
\usepackage[colorlinks=true, allcolors=blue]{hyperref}
\usepackage{url}            
\usepackage{booktabs}       
\usepackage{amsfonts}       
\usepackage{nicefrac}       
\usepackage{microtype}      
\usepackage{xcolor}         
\usepackage{multirow}

\usepackage{amsmath, amssymb, amsfonts, dsfont, mathabx,mathtools}

\usepackage{float}
\usepackage{wrapfig}
\usepackage{subcaption}
\usepackage{thmtools,thm-restate}
\usepackage{algorithmic}
\usepackage{algorithm}
\usepackage{graphicx}
\usepackage{caption, subcaption}
\usepackage{csvsimple}
\usepackage{mymacros}
\graphicspath{{./images/}}

\author{
  Wei-Chen Wang  \\
    \texttt{wangeric@mit.edu} \\
    Cleanlab, MIT 
    \And
    Jonas Mueller \\
    \texttt{jonas@cleanlab.ai} \\
    Cleanlab
}

\newcommand{\papertitle}{Detecting Label Errors in Token Classification Data}

\newcommand{\ssa}{\texttt{predicted-difference}} 
\newcommand{\ssb}{\texttt{bad-token-counts}}
\newcommand{\ssc}{\texttt{bad-token-counts-avg}}
\newcommand{\ssd}{\texttt{bad-token-counts-min}}
\newcommand{\sse}{\texttt{good-fraction}}
\newcommand{\ssf}{\texttt{penalize-bad-tokens}}
\newcommand{\ssg}{\texttt{average-quality}}
\newcommand{\ssh}{\texttt{product}}
\newcommand{\ssi}{\texttt{expected-bad}}
\newcommand{\ssj}{\texttt{expected-alt}}
\newcommand{\ourmethod}{\texttt{worst-token}} 
\newcommand{\varalt}{\texttt{worst-token-min-alt}}
\newcommand{\varsoft}{\texttt{worst-token-softmin}}

\newcommand{\tsa}{\texttt{sc}} 
\newcommand{\tsb}{\texttt{nm}} 
\newcommand{\tsc}{\texttt{cwe}}
\newcommand{\tq}[1]{\texttt{``#1''}}
\title{\papertitle{}}
\begin{document}
\title{\papertitle}
\maketitle
\thispagestyle{empty}

\begin{abstract}
Mislabeled examples are a common issue in real-world data, particularly for tasks like token classification where many labels must be chosen on a fine-grained basis. Here we consider the task of finding sentences that contain label errors in token classification datasets. We study 11 different straightforward methods that score tokens/sentences based on the predicted class probabilities output by a (any) token classification model (trained via any procedure). In precision-recall evaluations based on real-world label errors in entity recognition data from CoNLL-2003, we identify a simple and effective method that consistently detects those sentences containing label errors when applied with different token classification models. 

\end{abstract}

\section{Introduction}
\label{sec:intro}

It has recently come to light that many supervised learning datasets contain numerous incorrectly labeled examples \cite{northcutt2021labelerrors}. To efficiently improve the quality of such data,  Label Error Detection (LED) has emerged as a task of interest \cite{kuan22, northcutt2021confidentlearning, muller2019mislabeled}, in which algorithms flag examples whose labels are likely wrong for reviewers to inspect/correct. This may be done by means of a score for each example which reflects its estimated \emph{label quality}. A useful score ranks mislabeled examples higher than others, allowing reviewers to much more efficiently identify the label errors in a dataset.

This paper considers LED for token classification tasks (such as \emph{entity recognition}) in which each token in a sentence has been given its own class label. While it is possible to score the labels of individual tokens, reviewing a candidate token flagged as potentially mislabeled requires looking at the entire sentence to understand the broader context. Here we propose \ourmethod{}, a method to score sentences based on the likelihood that they contain some mislabeled tokens, such that sentences can be effectively ranked for efficient label review\footnote{Code to run our proposed method is available here: \url{https://github.com/cleanlab/cleanlab} \\ Code to reproduce our results:  \url{https://github.com/cleanlab/token-label-error-benchmarks}}. We evaluate the LED performance of this approach and others on real-world data with naturally occuring label errors, unlike many past LED evaluations based on synthetically-introduced label errors \cite{brodley1999identifying, muller2019mislabeled, gu2021instance}, for which conclusions may differ from real-world errors \cite{kuan22,jiang2020controlled,wei2022learning}.

\paragraph{Related Work.}
Extensive research has been conducted on standard classification with noisy labels \cite{song2022survey, wei2022learning, chen2019understanding, muller2019mislabeled, natarajan2013learning, brodley1999identifying}. Our work builds on label quality scoring methods for classification data studied by \citet{northcutt2021confidentlearning,kuan22}, which merely depend on predictions from a trained multiclass classification model. These methods are straightforward to implement and broadly applicable, being compatible with any classifier and training procedure, as is our \ourmethod{} method.
\looseness=-1

Only a few prior works have studied label error detection for token classification tasks specifically \cite{wang2019crossweigh,reiss2020identifying,klie2022annotation}. Also aiming to  score the overall label quality of an entire sentence like us, \citet{wang2019crossweigh} propose \texttt{CrossWeigh} which: trains a large ensemble of token classification models with \emph{entity-disjoint} cross-validation, and scores a sentence based on the number of ensemble-member class predictions that deviate from the given label, across all tokens in the sentence. \citet{reiss2020identifying} propose a similar approach to LED in token classification, which also relies on counting deviations between token labels and corresponding class predictions output by a large ensemble of diverse models trained via cross-validation. 
Unlike our \ourmethod{} method, the methods of \citet{wang2019crossweigh, reiss2020identifying}  are more computationally complex due to ensembling many models, and do not account for the confidence of individual predictions (as they are based on hard class predictions rather than estimated class probabilities). Given the LED performance of \ourmethod{} 
improves with a more accurate token classifier, we expect \ourmethod{} to benefit from ensembling's reliable predictive accuracy improvement in the same way that ensembling benefits the methods of \citet{wang2019crossweigh, reiss2020identifying}. We also tried entity-disjoint vs.\ standard data splitting, but did not observe benefits from the former variant (it often removes a significant number of entities).

\citet{klie2022annotation} study various approaches for LED in token classification, but only consider scores for individual tokens rather than entire sentences. Here we compare against some of the methods that performed best in their study. The studies of \citet{klie2022annotation, northcutt2021confidentlearning,kuan22} indicate that label errors can be more effectively detected by considering the confidence level of classifiers rather than only their hard class predictions. Instead depending on predicted class probabilities for each token, our \ourmethod{} method appropriately accounts for classifier confidence.
\looseness=-1

\section{Methods}
\label{sec:methods}
Typical token classification data is composed of many sentences (i.e.\ training instances), each of which is split into individual tokens (words or sub-words) where each token is labeled as one of $K$ classes (i.e. entities in entity recognition).
Given a sentence $x$, a trained token classification model $M(\cdot)$ outputs predicted probabilities $\mathbf{p} = M(x)$ where $p_{ij}$ is the probability that the $i$th token in sentence $x$  belongs to class $j$. Throughout, we assume these probabilities are \emph{out-of-sample}, computed from a copy of the model that did not see $x$ during training (e.g. because $x$ is in test set, or cross-validation was utilized).

Using $p$, we first consider evaluating the individual per-token labels. Here we apply effective LED methods for standard classification settings \cite{northcutt2021confidentlearning} by simply treating each token as a separate independent instance (ignoring which sentence it belongs to). Following \citet{kuan22}, we compute a label quality score $q_i \in [0,1]$ for the $i$th token (assume it is labeled as class $k$) via one of the following options:
\begin{itemize}[leftmargin=*]
    \item self-confidence (\tsa): $q_i = p_{ik}$, \ i.e. predicted probability of the given label for this token.
    \item normalized margin (\tsb): $q_i = p_{ik} - p_{i \tilde{k}}$ with $\tilde{k} = \argmax_j \ \{ p_{ij} \} $
    \item confidence-weighted entropy (\tsc): $ \displaystyle q_i = \frac{p_{ik}}{H(p_i)}$ \ where $\displaystyle H(p_i) = - \frac{1}{\log K} \sum_{j=1}^K p_{ij} \log (p_{ij}) $
\end{itemize}
Higher values of these label quality scores correspond to tokens whose label is more likely to be correct \cite{kuan22}. We can alternatively evaluate the per-token labels via the Confident Learning algorithm of \citet{northcutt2021confidentlearning}, which classifies each token as correclty labeled or not ($b_i = 1$ if this token is flagged as likely mislabeled, = 0 otherwise) based on adaptive thresholds set with respect to per-class classifier confidence levels.

For one sentence with $n$ word-level tokens, we thus have: 
\begin{itemize}[leftmargin=*]
    \item $\mathbf{p}$, a $n\times K$ matrix where $p_{ij}$ is model predicted probability that the $i$th token belongs to class $j$. 
    \item $\mathbf{l}=[l_1,\dots,l_n]$, where $l_i\in\{0,\dots,K-1\}$ is the given class label of the $i$th token. 
    \item $\mathbf{q}=[q_1,\dots,q_n]$, where $q_i$ is a label quality score for the $i$th token (one of the above options).
    \item $\mathbf{b}=[b_1,\dots,b_n]$, where $b_i=1$ if $i$th token is flagged as potentially mislabeled, otherwise $b_i=0$.
    \looseness=-1
\end{itemize}

Recall that to properly verify whether a token is really mislabeled, a reviewer must read the full sentence containing this token to understand the broader context. Thus the most efficient way to review labels in a dataset is to prioritize inspection of those \emph{sentences most likely to contain a mislabeled token}. 
We consider $11$ methods to estimate an overall quality score $s(x)$ for the sentence $x$, where higher values correspond to sentences whose labels are more likely all correct. 
\begin{enumerate}
    \item \ssa: The number of disagreements between the given and model-predicted class labels over the tokens in the sentence, also utilized in the methods of  \citet{wang2019crossweigh,reiss2020identifying}. Here we 
    break sentence-score ties in favor of the highest-confidence disagreement. More formally: 
    $$s(x) = -\lvert\mathcal{R}\rvert-\max_{i\in\mathcal{R}}p_{i, \widehat{l}_i}$$ 
    where $\widehat{l}_i=\argmax_j \{p_{ij}\}$ and  $\mathcal{R}=\{i:\widehat{l}_i \neq l_i\}$. If $\mathcal{R}=\emptyset$, we let $\max_{i\in\mathcal{R}}p_{i,\widehat{l}_i}=0$. 
    \item \ssb: $ s(x) = - \sum_i b_i$, the number of Confident Learning flagged tokens. Similarly considered by \citet{klie2022annotation}, this approach is a natural token-classification extension of the method of \citet{northcutt2021confidentlearning} for LED in standard classification tasks.
    \item \ssc: Again scoring based on number of tokens flagged as potentially mislabeled, but now breaking ties primarily via the average label quality score of the flagged tokens and secondarily via the average label quality score of the other tokens. More formally: 
    $$s(x) = -\sum_i b_i + \frac{1}{\lvert \mathcal{R} \rvert}\sum_{i \in  \mathcal{R}} q_i+ \frac{\epsilon}{\lvert \mathcal{S} \rvert}\sum_{i\in \mathcal{S}}q_i$$ 
    where $\mathcal{R}=\{i:b_i=1\}$, $\mathcal{S}=\{i:b_i=0\}$, and $\epsilon$ is some small constant. 
    \item \ssd: Similar to \ssc, but break ties using minimum token quality rather than average token quality. More formally: 
    $$s(x) = -\sum_i b_i  + \min_{i\in \mathcal{R}} q_i+\epsilon\cdot\min_{i\in \mathcal{S}}q_i$$ 
    \item \sse: Fraction of tokens not flagged as potential issues, $\displaystyle s(x) = - \frac{1}{n} \sum_{i=1}^n b_i$. 
    \item \ssf: Penalize flagged tokens based on their corresponding label quality scores. More formally, 
    $$s(x) =1-\frac{1}{n}\sum_{i=1}^nb_i(1-q_i)$$ 
    \item \ssg: Average label quality of tokens in the sentence, $\displaystyle s(x) =  \frac{1}{n} \sum_{i=1}^n q_i$. 
    \item \ssh: $ s(x) = \sum_i \log(q_i+c)$, where $c$ is a constant hyperparameter. This score places greater emphasis on tokens with low estimated label-quality, while still being influenced by all tokens' quality (like the previous  \ssg{} method). With $q$ based on \tsa{} or \tsb{} token-scores, the \ssh{} and \ssg{} methods are natural sentence extensions of the CU or PM methods considered in \citet{klie2022annotation} for token-level LED.

    \item \ssi: A rough approximation of the expected number of mislabeled tokens in sentence. More formally: $$s(x) = \sum_{j=1}^{\min(n,J)}j\cdot q^{(j)}$$ 
    where $q^{(i)}$ is the $i$th lowest token label quality score in this sentence, and $J$ is a hyperparameter. If using the \tsa{} label-quality score, $1-q^{(i)}$ can be considered a loose proxy for the probability of having at least $i$ label errors in this sentence. 
    \item \ssj: Similar to \ssi, but only considering the likelihood of any label error rather than how many might be in this sentence. More formally: $$s(x) = \sum_{j=1}^{\min(n,J)}q^{(j)}$$ 
    \item \ourmethod: The quality of the worst-labeled token in the sentence determines its overall quality score, $\displaystyle s(x) =  \min \{ q_1, q_2, \dots, q_n\}$. This is a reasonable way to rank the sentences most likely to have some label error, i.e. those most worthy of manual review. 
\end{enumerate}

 \section{Results}
\label{sec:experiments} 

For evaluation, we apply each sentence scoring method to the given class labels in the CoNLL-2003 named entity recognition dataset \cite{conll-dataset}. We restrict our attention to the test set, for which all ground truth label errors were identified by \citet{wang2019crossweigh} (see Appendix \ref{sec:improved}).
We consider two different models to produce per-token predicted probabilities: \texttt{bert} \cite{original-bert-paper} and \texttt{xlm} \cite{original-xlm-paper}, and we consider a second variant of the dataset (\texttt{unmerged}) with more classes based on additional consideration of \texttt{B-} and \texttt{I-} entity-prefixes.
See Appendix \ref{sec:details} for all details.

Recall the sentence scores $s(x)$ are used to prioritize which sentences most likely contain label errors. Thus we consider evaluation metrics from information retrieval, which depend on the ranking of sentences induced by $s(x)$ rather than the magnitude of its values. Sentences that contain any mislabeled token are considered true positives when we compute metrics like: \textbf{AUROC} (and \textbf{AUPRC}) for area under the receiver operating characteristic (and precision-recall) curve. 
Our third metric, \textbf{Lift @ \#Errors}, measures how many times more prevalent labels errors are within the top-$T$ scoring sentences vs.\ all sentences. Here $T$ is the number of true positives ($T=184$ for \texttt{bert} and \texttt{xlm},  $T=186$ for \texttt{bert-unmerged}). 
The Lift metric favors high-precision scores, while AUROC and AUPRC consider both precision and recall, favoring scores capable of detecting a meaningful fraction of all true positives. AUPRC is sometimes preferred over AUROC in settings where  true positives are rare \cite{davis2006relationship, saito2015precision}. \looseness=-1

Table \ref {tab:auprc_main} presents some results and others are in Appendix \ref{sec:additionalresults}. Our results show that \ourmethod\space (using the \tsa\space token-score) generally achieves the best LED performance across the three experiments. 
To most usefully rank sentences for identifying label errors, one should thus account for classifier confidence but not be directly influenced by all tokens' estimated quality (which may be noisy). \looseness=-1


\begin{table}[h!]
\caption{AUPRC achieved by different sentence and token scoring methods.}
\label{tab:auprc_main}
\footnotesize 
\begin{center} 
\begin{tabular}{p{3.5cm} l p{2cm} p{2cm} p{2.1cm}}
\toprule 
\multicolumn{1}{l}{\textbf{Sentence Score} }  & \textbf{Token Score}  & \textbf{bert} & \textbf{xlm} & \textbf{bert-unmerged} \\  \midrule
\ssa                    &           & 0.3422 & 0.3412 & 0.3190 \\ \midrule
\ssb                    &           & 0.3087 & 0.3186 & 0.3291 \\ \midrule
\multirow{3}{*}{\ssc}   & \tsa      & 0.3740 & 0.3697 & 0.3768 \\
                        & \tsb      & 0.3702 & 0.3603 & 0.3740 \\
                        & \tsc      & 0.3597 & 0.3597 & 0.3609 \\ \midrule
\multirow{3}{*}{\ssd}   & \tsa      & 0.3804 & 0.3759 & 0.3901 \\
                        & \tsb      & 0.3744 & 0.3662 & 0.3822 \\
                        & \tsc      & 0.3695 & 0.3602 & 0.3607 \\ \midrule
\sse                    &           & 0.3131 & 0.3159 & 0.2996 \\ \midrule
\multirow{3}{*}{\ssg}   & \tsa      & 0.3022 & 0.3349 & 0.2574 \\
                        & \tsb      & 0.3066 & 0.3143 & 0.2648 \\
                        & \tsc      & 0.2767 & 0.3495 & 0.2572 \\ \midrule
\multirow{3}{*}{\ssf}   & \tsa      & 0.3423 & 0.3321 & 0.3229 \\
                        & \tsb      & 0.3368 & 0.3126 & 0.3221 \\
                        & \tsc      & 0.3191 & 0.3380 & 0.3023 \\ \midrule
\multirow{3}{*}{\ssh}   & \tsa      & 0.3794 & 0.3559 & 0.3726 \\
                        & \tsb      & 0.3807 & 0.3533 & 0.3823 \\
                        & \tsc      & 0.3519 & 0.3783 & 0.3359 \\ \midrule
\multirow{3}{*}{\ssi}   & \tsa      & 0.3383 & 0.3485 & 0.3532 \\
                        & \tsb      & 0.3776 & 0.3227 & 0.3513 \\
                        & \tsc      & 0.3191 & 0.3541 & 0.2980 \\ \midrule
\multirow{3}{*}{\ssj}   & \tsa      & 0.3927 & 0.3628 & 0.3614 \\
                        & \tsb      & 0.3850 & 0.3342 & 0.3603 \\
                        & \tsc      & 0.3335 & 0.3620 & 0.3114 \\ \midrule
\multirow{3}{*}{\ourmethod}   & \tsa      & \textbf{0.4357} & \textbf{0.4021} & \textbf{0.4236} \\
                        & \tsb      & 0.4243 & 0.3963 & 0.3933 \\
                        & \tsc      & 0.3215 & 0.3815 & 0.2974 \\ \midrule
\end{tabular}
\end{center} 
\end{table}

\clearpage
\bibliographystyle{abbrvnat}
\bibliography{tokenclassification}


\clearpage
\appendix

\begin{center}
{\bf \LARGE Appendix: \ 
\papertitle{}
}
\end{center}

\section{Experiment Details} 
\label{sec:details} 

\paragraph{Dataset.} The CoNLL-2003 dataset contains $4$ types of named entities: \texttt{PER} for persons, \texttt{ORG} for organizations, \texttt{LOC} for locations, \texttt{MISC} for miscellaneous other entities, with \texttt{O} being reserved as a label for other types of words that are not named entities. The dataset is in IOB2 format \cite{ratnaparkhi1998maximum}, such that all named entities possess an extra \texttt{B-} or \texttt{I-} prefix, which indicates whether this token is the Beginning of an entity or an Intermediate part of one. 
We consider this set of 9 classes in our \texttt{unmerged} version of the dataset. In the alternative version of the dataset, we ignore the \texttt{B-} and \texttt{I-} prefixes, and only consider the 5 more meaningful classes based on the different entity types.
In either setting, a sentence is considered ``mislabeled'' if it contains any word-level token whose given label in the original dataset does not match the ground truth label from a corrected version of the dataset.

\paragraph{Models.} For producing predicted probabilities to input into our label quality scoring methods, all models were only trained on the training set of CoNLL-2003, and we use them to produce held-out predictions for the test set. No model has access to the test set examples during training, nor any ground-truth label errors at any point in our evaluation.

We consider three different settings in our experiments, described below in the corresponding order:  \texttt{bert-unmerged}, \texttt{bert}, \texttt{xlm}. 
The first model we consider is a pre-trained \texttt{bert-base-NER} Transformer network, which has been found to be an effective token classifier for CoNLL-2003 including the \texttt{B-} and \texttt{I-} entity-prefixes as additional classes (for a total of 9 classes)  \cite{dslim}. 
We conduct two additional experiments to verify whether the label error detection results with this model remain consistent in other  settings. First, we omit the \texttt{B-} and \texttt{I-} prefixes such that we only focus on more severe error types, such as \texttt{LOC} vs. \texttt{ORG} rather than \texttt{B-LOC} vs. \texttt{I-LOC} label substitutions (which may be of less interest). With this reduced set of entities, we still use the same Bert network, which now predicts amongst a fewer set of 5 classes for each token.
Finally, to examine how our methods work when applied with different models, we also obtain a different set of model-predicted probabilities using another pre-trained XLM network: \texttt{xlm-roberta-large-finetuned-conll03-english} \cite{xlmhf}. Given that the model outputs predictions in the IOB format \cite{ramshaw1999text}, we again consider the reduced set of 5 classes. 

\paragraph{Data Processing.} Before applying our label error detection methods, we first preprocess the original CoNLL-2003 dataset. Sentences less than or equal to $1$ character, and sentences containing \tq{\#} are excluded. The latter because \tq{\#} is a special character reserved by many token classification models to represent subword tokens. We construct each sentence by joining the tokens separated by space, and perform some minor cleanup to ensure that the sentence follows writing conventions (such as no space before comma, and no space after open parenthesis). We convert all-caps tokens into lowercase except for the first character (i.e.\  \texttt{JAPAN} -> \texttt{Japan}), because token classification models tend to partition all-caps tokens into multiple subword-level tokens. This can sometimes result in some undesirable behaviors, such as converting \texttt{USA} to \texttt{Usa}, or \texttt{NBA} to \texttt{Nba}. We believe that the benefits outweigh the costs for CoNLL-2003, due to the prevalence of article headlines in which most, if not all, tokens in the sentence are all-caps. 

Next, we use a pre-trained token classification model to obtain the model-predicted probabilities. 
Modern Transformer models first tokenize the sentence into multiple subword-level tokens, possibly different from the given tokens individually labeled in the original dataset. These ``subword-level'' tokens are typically smaller units than word-level tokens, and the trained model outputs a probability distribution over possible classes for each such token. 

Here we reduce these probabilities from subword-level to word-level tokens, so that we can evaluate the given labels as outlined in the main text. Consider the sentence: 

\quad\texttt{Minnesota Timberwolves (MIN)} 

where the given tokens from original dataset are \texttt{[\tq{Minnesota}, \tq{Timberwolves}, \tq{(}, \tq{MIN}, \tq{)}]}, and the model tokenizes the sentence into subword-level tokens: \texttt{[\tq{Minnesota}, \tq{Timber}, \tq{wolves}, \tq{(MIN)}]}\footnote{Different models may result in different tokenization.}. Let $\mathbf{p}^{(i)}=[p^{(i)}_1, p^{(i)}_2, \dots, p^{(i)}_K]$ denote the model-predicted probabilities of the $i$th subword-level token, where $K$ is the number of possible classes. To obtain the predicted probability distribution for \tq{Timberwolves} (used to evaluate its given label in the original dataset), we take the average of the (model-estimated) probabilities for \tq{Timber} $\mathbf{p}^{(2)}$, and \tq{wolves} $\mathbf{p}^{(3)}$. We assign the probability for \tq{(MIN)} $\mathbf{p}^{(4)}$ directly to \tq{(}, \tq{MIN} and \tq{)}. Similar strategies are applied for all other model-tokenization-induced discrepancies in the data.

We also experimented with alternative probability-pooling methods for adjacent subword-level tokens, such as: a weighted average (with weights proportional to the number of characters in each subword-level token, such that longer strings receive a higher weight), or considering only the predicted probabilities for first subword-level token. Both of these variants produced little differences in their LED performance from the average-pooling technique we employ for the results shown in this paper.

\paragraph{Hyperparameter Settings.}  5 of the sentence scoring methods considered in this paper contain a hyperparameter (but not our proposed \ourmethod{} method).  To ensure these other methods are not disadvantaged vs.\ \ourmethod{}, we tried many different values of their hyperparameter and report results for the best hyperparameter value. See Appendix \ref{sec:resultsforvariants} for descriptions of some additional methods not introduced in the main text.  The values considered for each method are as follows, with the selected hyperparameter in \textbf{bold}. 
\begin{itemize}
    \item \ssh: $10^{-1}, 10^{-1.5}, 10^{-2}, 10^{-2.5}, \mathbf{10^{-3}}$
    \item \varalt: $0.01, 0.02, 0.03, 0.04, 0.05, 0.06, 0.07, 0.08, 0.09, \mathbf{0.1}$ 
    \item \varsoft: $10^{-1}, 10^{-1.1}, 10^{-1.2}, 10^{-1.3}, 10^{-1.4}, \mathbf{10^{-1.5}}$
    \item \ssi: $\mathbf{2}, 3, 4$ 
    \item \ssj: $\mathbf{2}, 3, 4$
\end{itemize}
Note that \ssi{} and \ssj{} with a hyperparameter value of 1 are identical to \ourmethod{}, so this value is excluded from our hyperparameter sweep.

\clearpage
\section{Ground Truth Label Errors in CoNLL (via CoNLL++)}
\label{sec:improved}

\citet{wang2019crossweigh} manually corrected the entire test set of CoNLL-2003, discovering $186$ sentences ($5.38\%$ of the original data) that contain at least one token label error. Through extensive manual inspection, we comprehensively verified  that basically all CoNLL-2003 label errors (in test set) have been fixed by \citet{wang2019crossweigh}, and their proposed corrections are reliable (we did not find false positives or negatives). \citet{wang2019crossweigh} named their corrected version \texttt{CoNLL++}, which we consider as a source of ground truth to validate candidate label errors for this study. 

\citet{reiss2020identifying} also proposed a corrected set of labels for CoNLL-2003, identified via semi-supervised algorithms and limited manual label verification.
However, close examination of their proposed corrections reveals that many of them are fundamentally incorrect and many real CoNLL label errors (found by \citet{wang2019crossweigh}) were not properly identified/fixed by \citet{reiss2020identifying}. For example, consider the first sentence from the test set: 

\quad\texttt{Soccer - Japan get lucky win, China in surprise defeat.} 

In the original CoNLL-2003 dataset, \texttt{Japan} is labeled \texttt{LOC}, and \texttt{China} is labeled \texttt{PER} (this happens to be a label error, it should be \texttt{LOC} instead). Here, we ignore the \texttt{B-} and \texttt{I-} prefix. In the dataset corrected by \citet{reiss2020identifying}, \texttt{Japan} and \texttt{China} are instead labeled \texttt{ORG}. In addition, consider another sentence from the test set: 

\quad\texttt{...is not for a stronger dollar either," said Sumitomo's Note.} 

In the dataset corrected by \citet{reiss2020identifying}, \tq{Note} is labeled as \texttt{PER}, while the correct label should be \texttt{O}. The token is also labeled incorrectly in the original dataset, but is corrected by \citet{wang2019crossweigh}. Overall comparing against the high-quality (comprehensively manually verified) label corrections in \texttt{CONLL++}, we found the (mostly algorithmically)  CoNLL-corrected dataset from \citet{reiss2020identifying} contains many corrections which are not actually valid (we estimate around $8\%$ of their proposed corrections are wrong). Hence, we opted not to base any evaluations on this dataset from \citet{reiss2020identifying}. 
This highlights the difficulty of algorithmically correcting an entire dataset's labels, which is why we focus on label error detection in this work, developing methods that enable human reviewers to quickly find and fix the label errors.

\begin{table}[h]
\centering 
\caption{Label noise distribution in the classes of the (\texttt{unmerged}) CoNLL-2003 dataset, calcuated by comparing original labels against the corrected versions in CoNLL++. The cell in the $i$th row and $j$th column lists the percentage of tokens that are labeled class $i$ in CoNLL++ (ground truth label) and mislabeled as class $j$ in the original CoNLL-2003 dataset. }
\label{tab:noisematrix}
\vspace*{0.7em}
\begin{footnotesize}
\begin{tabular}{l|rrrrrrrrr}
       & O       & B-MISC & I-MISC & B-PER  & I-PER  & B-ORG  & I-ORG  & B-LOC  & I-LOC  \\ \hline
O      & -       & 0.01\% & 0.02\% & 0.01\% &        & 0.01\% &        &        &        \\
B-MISC & 5.39\%  & -      &        &        &        & 0.14\% &        & 2.49\% &        \\
I-MISC & 18.90\% & 1.57\% & -      &        &        &        &        &        &        \\
B-PER  & 0.49\%  & 0.12\% &        & -      &        & 0.06\% &        & 0.19\% &        \\
I-PER  & 0.17\%  &        &        & 0.43\% & -      &        &        &        & 0.09\% \\
B-ORG  & 0.82\%  & 1.17\% &        & 0.18\% &        & -      &        & 1.87\% &        \\
I-ORG  & 3.18\%  &        & 0.68\% &        & 0.34\% & 0.34\% & -      & 0.23\% & 0.80\% \\
B-LOC  & 0.91\%  & 0.49\% &        & 0.18\% &        & 0.43\% & 0.06\% & -      &        \\
I-LOC  & 2.70\%  &        & 0.77\% &        &        &        & 0.39\% &        & -     
\end{tabular}
\end{footnotesize}
\end{table}

\clearpage
\section{Additional Results} 
\label{sec:additionalresults} 
Tables \ref{lift_table} and \ref{auroc_table} provide additional sentence scoring results for the Lift and AUROC metrics. 
Note the token score field is left empty for sentence scoring methods that do not rely on token scores. 

\begin{table}[htp]
\caption{Lift @ \#Errors for varying sentence and token scoring methods.} 
\label{lift_table}
\footnotesize 
\begin{center} 
\begin{tabular}{p{3.5cm} l p{2cm} p{2cm} p{2.1cm}}
\midrule \\ [-1.8ex] 
\multicolumn{1}{l}{Sentence score}  & Token Score  & bert & xlm & bert-unmerged \\ [0.2ex] \midrule
\ssa                    &           & 7.19 & 6.59 & 7.14 \\ \midrule 
\ssb                    &           & 7.30 & 6.89 & 8.03 \\ \midrule 
\multirow{3}{*}{\ssc}   & \tsa      & 7.80 & 7.09 & 8.13 \\
                        & \tsb      & 7.60 & 7.40 & 8.13 \\
                        & \tsc      & 7.30 & 6.89 & 7.24 \\ \midrule
\multirow{3}{*}{\ssd}   & \tsa      & 7.80 & 7.09 & 8.23 \\
                        & \tsb      & 7.60 & 7.40 & 8.13 \\
                        & \tsc      & 7.30 & 6.89 & 7.24 \\ \midrule
\sse                    &           & 5.88 & 6.18 & 5.45 \\ \midrule 
\multirow{3}{*}{\ssf}   & \tsa      & 5.98 & 6.18 & 5.55 \\
                        & \tsb      & 5.98 & 6.18 & 5.65 \\
                        & \tsc      & 6.38 & 6.08 & 5.75 \\ \midrule
\multirow{3}{*}{\ssg}   & \tsa      & 5.07 & 5.88 & 4.76 \\
                        & \tsb      & 5.07 & 5.98 & 4.66 \\
                        & \tsc      & 5.37 & 6.18 & 5.26 \\ \midrule
\multirow{3}{*}{\ssh}   & \tsa      & 7.30 & 6.99 & 7.24 \\
                        & \tsb      & 7.60 & 7.40 & 7.64 \\
                        & \tsc      & 6.79 & 7.09 & 7.24 \\ \midrule

\multirow{3}{*}{\ssi}   & \tsa      & 6.99 & 6.48 & 6.84 \\
                        & \tsb      & 6.69 & 6.48 & 7.24 \\
                        & \tsc      & 6.59 & 6.08 & 7.34 \\ \midrule
\multirow{3}{*}{\ssj}   & \tsa      & 6.99 & 6.48 & 6.94 \\
                        & \tsb      & 6.89 & 6.59 & 7.24 \\
                        & \tsc      & 7.09 & 6.89 & 7.34 \\ \midrule
\multirow{3}{*}{\ourmethod}   & \tsa      & \textbf{9.02} & \textbf{8.71} & \textbf{8.83} \\
                        & \tsb      & 9.02 & 8.71 & 8.73 \\
                        & \tsc      & 7.40 & 7.90 & 6.35 \\ \midrule
\end{tabular}
\end{center} 
\end{table}

\begin{table}[h]
\caption{AUROC for varying sentence and token scoring methods.} 
\label{auroc_table}
\footnotesize 
\begin{center} 
\begin{tabular}{p{3.5cm} l p{2cm} p{2cm} p{2.1cm}}
\midrule \\ [-1.8ex] 
\multicolumn{1}{l}{Sentence score}  & Token Score  & bert & xlm & bert-unmerged \\ [0.2ex] \midrule
\ssa                    &           & 0.8639 & 0.8633 & 0.8559 \\ \midrule 
\ssb                    &           & 0.8188 & 0.8421 & 0.8114 \\ \midrule 
\multirow{3}{*}{\ssc}   & \tsa      & 0.8934 & 0.9033 & 0.8718 \\
                        & \tsb      & 0.8932 & 0.9030 & 0.8702 \\
                        & \tsc      & 0.8902 & 0.9012 & 0.8725 \\ \midrule
\multirow{3}{*}{\ssd}   & \tsa      & 0.9026 & 0.9106 & 0.8885 \\
                        & \tsb      & 0.9025 & 0.9102 & 0.8854 \\
                        & \tsc      & 0.9023 & 0.9091 & 0.8892 \\ \midrule
\sse                    &           & 0.8147 & 0.8393 & 0.8049 \\ \midrule 
\multirow{3}{*}{\ssf}   & \tsa      & 0.8151 & 0.8396 & 0.8053 \\
                        & \tsb      & 0.8151 & 0.8396 & 0.8058 \\
                        & \tsc      & 0.8162 & 0.8401 & 0.8064 \\ \midrule
\multirow{3}{*}{\ssg}   & \tsa      & 0.8553 & 0.8895 & 0.8079 \\
                        & \tsb      & 0.8560 & 0.8894 & 0.8068 \\
                        & \tsc      & 0.8578 & 0.8913 & 0.8305 \\ \midrule
\multirow{3}{*}{\ssh}   & \tsa      & 0.8900 & 0.8647 & 0.8811 \\
                        & \tsb      & 0.8905 & 0.8646 & 0.8784 \\
                        & \tsc      & 0.8870 & 0.8683 & 0.8783 \\ \midrule
\multirow{3}{*}{\ssi}   & \tsa      & 0.8946 & 0.9026 & 0.8724 \\
                        & \tsb      & 0.8948 & 0.9017 & 0.8705 \\
                        & \tsc      & 0.8922 & 0.9033 & 0.8778 \\ \midrule
\multirow{3}{*}{\ssj}   & \tsa      & 0.8963 & 0.9060 & 0.8776 \\
                        & \tsb      & 0.8972 & 0.9044 & 0.8759 \\
                        & \tsc      & 0.8950 & 0.9048 & 0.8825 \\ \midrule
\multirow{3}{*}{\ourmethod}   & \tsa      & 0.9058 & \textbf{0.9141} & \textbf{0.8905} \\
                        & \tsb      & \textbf{0.9059} & 0.9134 & 0.8852 \\
                        & \tsc      & 0.8996 & 0.9121 & 0.8834 \\ \midrule
\end{tabular}
\end{center} 
\end{table}
\clearpage 

\begin{figure}[h]
\begin{center}
\textbf{(a)} \text{bert} \\
 \vspace{-0.8cm} 
\subfloat{%
  \includegraphics[clip,width=0.92\textwidth]{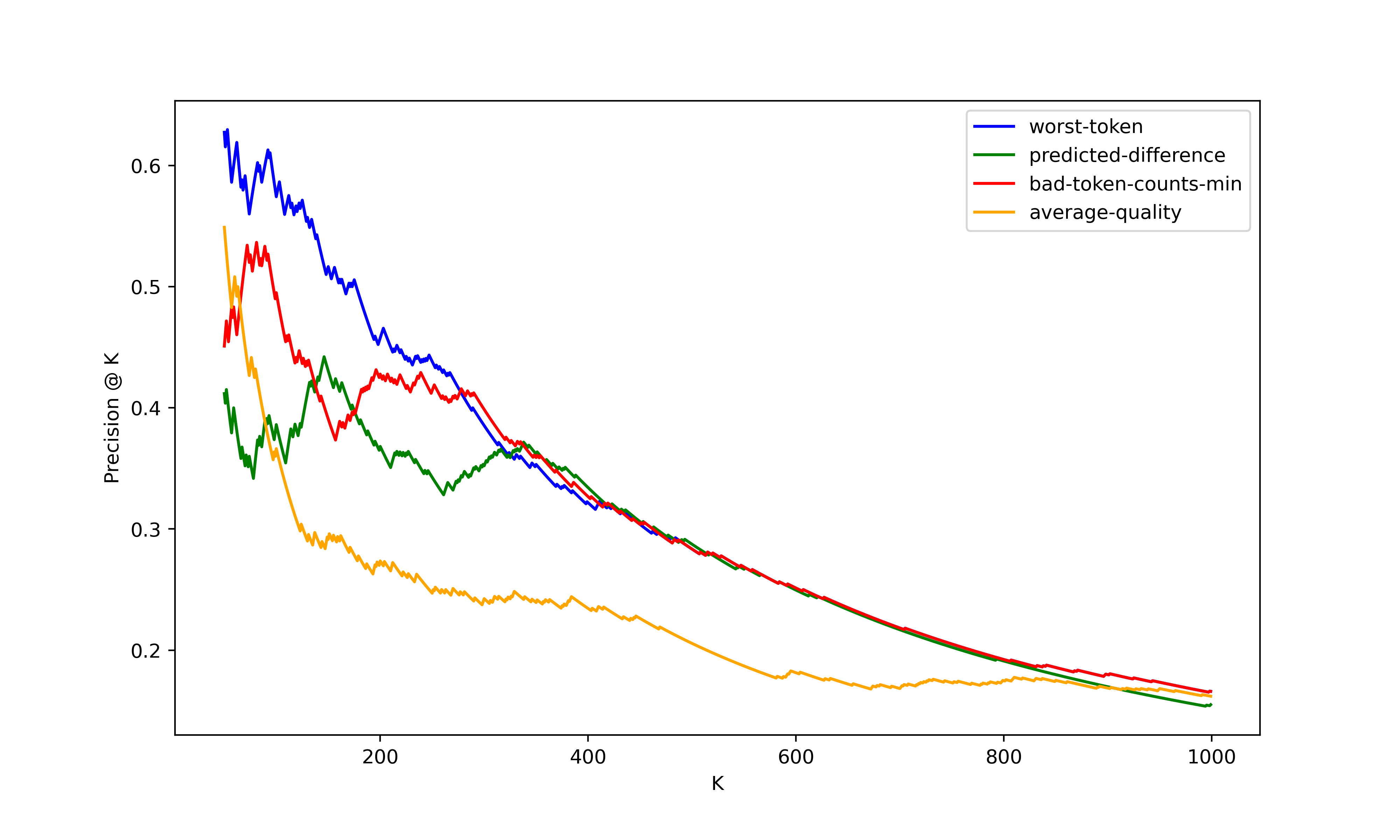}%
}
 \vspace{-0.3cm} 
\\ \textbf{(b)} \texttt{xlm} \\
 \vspace{-0.7cm} 
\subfloat{%
  \includegraphics[clip,width=0.92\textwidth]{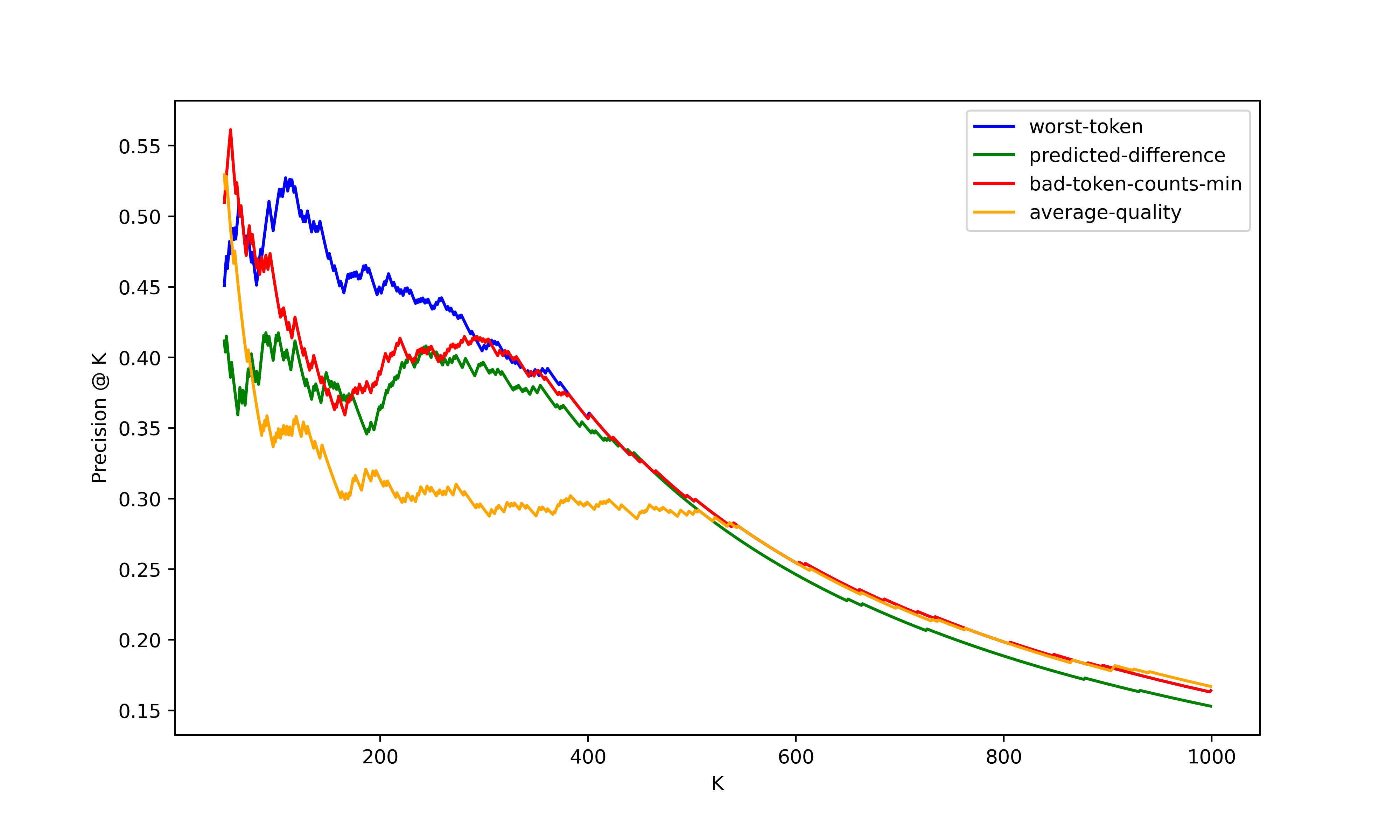}%
}
 \vspace{-0.3cm} 
\\ \textbf{(c)} \texttt{bert-unmerged} \\
 \vspace{-0.8cm} 
\subfloat{%
  \includegraphics[clip,width=0.92\textwidth]{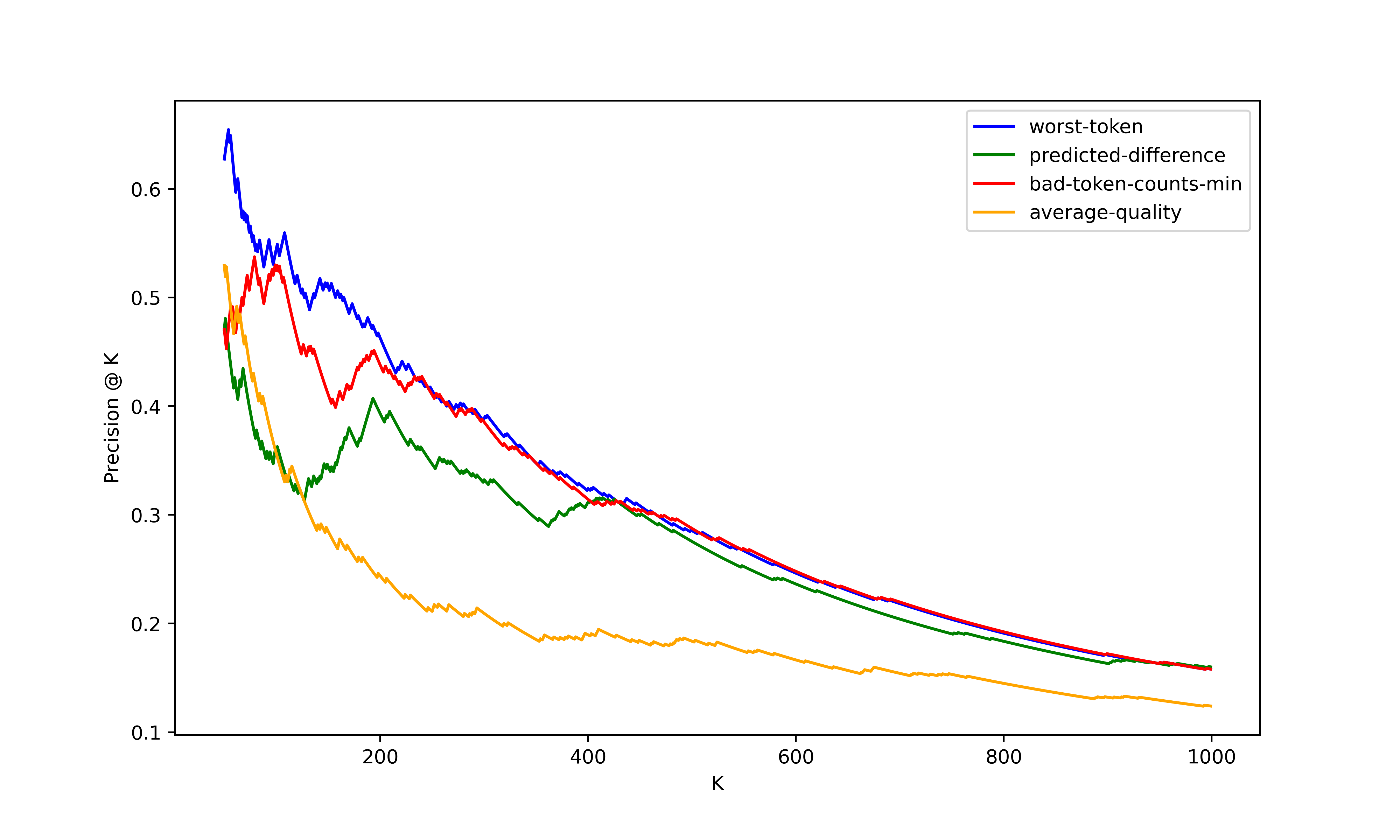}%
}
\vspace{-0.4cm} 
\end{center}
\caption{Precision @ $K$ for detecting sentences that contain a label error via our \ourmethod{} sentence scoring method and three others (over different values of $K$).
Across the 3 experiments, \ourmethod{} consistently detects sentences containing label errors with higher overall precision.} 
\end{figure}
\clearpage 

\section{Variants of our \ourmethod{} method} 
\label{sec:resultsforvariants}
For completeness, we also study some minor variants of our proposed methodology. Using the notation in Section \ref{sec:methods}, we consider the following alternative methods to produce a sentence score $s(x)$ for sentence $x$:
\begin{itemize}[leftmargin=*]
    \item \varalt: We add quality-score penalty $d$ for tokens flagged as likely label errors by Confident Learning \cite{northcutt2021confidentlearning}, and then consider the worst token based on the penalized quality scores. More formally:
    $$ s(x) =\min_i(q_i+d \cdot b_i)$$ 
    Recall $i$ ranges over the tokens in sentence $x$. We experimented with different  values of constant hyperparameter $d$. When $d \geq 1$, sentences that contain at least one token flagged as a likely label error by Confident Learning are completely separated from the remaining sentences. In other words, if a sentence does not include any flagged tokens, it is guaranteed to rank below all of the sentences with at least one flagged token. 
    \item \varsoft: Instead of considering only the worst token's quality score, we softly consider all other token's quality-score (to a lesser degree) as well in the overall sentence score.  More formally:
    $$ s(x) =\langle\mathbf{q},\text{softmax}_t(\mathbf{1}-\mathbf{q})\rangle$$ 
    Here $\langle\cdot,\cdot\rangle$ denotes the inner dot product. The temperature of the softmax $t$ is a constant hyperparameter we tried different values for. Settings of $t$ closer to 0 make this approach converge to our \ourmethod{} method, whereas this approach converges to the \ssg{} method with large values of $t$.
\end{itemize}

We evaluate the Lift, AUROC and AUPRC of our proposed method and its variants for comparison. The setup is the same as before, where we report results from the variants with the best hyperparameter settings we could find (to ensure they are not unfairly disadvantaged against \ourmethod{}). These settings were $d = 0.1$ for \varalt{} and $t = 10^{-1.5}$ for \varsoft{}. 
The best performing scoring method is highlighted in \textbf{bold}. 
\begin{table}[htp]
\caption{Lift @ \#Errors for our proposed method \ourmethod\space and its variants.} 
\footnotesize 
\label{var:lift} 
\begin{center} 
\begin{tabular}{p{3.5cm} l p{2cm} p{2cm} p{2.1cm}}
\midrule \\ [-1.8ex] 
\multicolumn{1}{l}{Sentence score}  & Token Score  & bert & xlm & bert-unmerged \\ [0.2ex] \midrule
\multirow{3}{*}{\ourmethod}   & \tsa      & 9.02 & 8.71 & 8.83 \\
                        & \tsb      & 9.02 & 8.71 & 8.73 \\
                        & \tsc      & 7.40 & 7.90 & 6.35 \\ \midrule
\multirow{3}{*}{\varalt}   & \tsa      & 9.02 & 8.71 & 8.63 \\
                        & \tsb      & 9.02 & 8.71 & 8.73 \\
                        & \tsc      & 7.90 & 7.70 & 7.83 \\ \midrule
\multirow{3}{*}{\varsoft}   & \tsa      & 8.92 & \textbf{8.82} & 8.53 \\
                        & \tsb      & \textbf{9.83} & 8.51 & \textbf{9.12} \\
                        & \tsc      & 7.50 & 8.00 & 6.35 \\ \midrule
\end{tabular}
\end{center} 
\end{table} 

\begin{table}[htp]
\caption{AUPRC for our proposed method \ourmethod\space and its variants.} 
\footnotesize 
\label{var:auprc} 
\begin{center} 
\begin{tabular}{p{3.5cm} l p{2cm} p{2cm} p{2.1cm}}
\midrule \\ [-1.8ex] 
\multicolumn{1}{l}{Sentence score}  & Token Score  & bert & xlm & bert-unmerged \\ [0.2ex] \midrule
\multirow{3}{*}{\ourmethod}   & \tsa      & 0.4357 & 0.4021 & \textbf{0.4236} \\
                        & \tsb      & 0.4243 & 0.3963 & 0.3933 \\
                        & \tsc      & 0.3215 & 0.3815 & 0.2974 \\ \midrule
\multirow{3}{*}{\varalt}   & \tsa      & \textbf{0.4408} & 0.4019 & 0.4181 \\
                        & \tsb      & 0.4238 & 0.3965 & 0.3945 \\
                        & \tsc      & 0.3643 & 0.3819 & 0.3081 \\ \midrule
\multirow{3}{*}{\varsoft}   & \tsa      & 0.4402 & \textbf{0.4063} & 0.4171 \\
                        & \tsb      & 0.4388 & 0.3980 & 0.4065 \\
                        & \tsc      & 0.3185 & 0.3799 & 0.2937 \\ \midrule
\end{tabular}
\end{center} 
\end{table}
\begin{table}[htp]
\caption{AUROC for our proposed method \ourmethod\space and its variants.} 
\footnotesize 
\label{var:auroc} 
\begin{center} 
\begin{tabular}{p{3.5cm} l p{2cm} p{2cm} p{2.1cm}}
\midrule \\ [-1.8ex] 
\multicolumn{1}{l}{Sentence score}  & Token Score  & bert & xlm & bert-unmerged \\ [0.2ex] \midrule
\multirow{3}{*}{\ourmethod}   & \tsa      & 0.9058 & \textbf{0.9141} & \textbf{0.8905} \\
                        & \tsb      & 0.9059 & 0.9134 & 0.8852 \\
                        & \tsc      & 0.8996 & 0.9121 & 0.8834 \\ \midrule
\multirow{3}{*}{\varalt}   & \tsa      & \textbf{0.9066} & 0.9140 & 0.8897 \\
                        & \tsb      & 0.9058 & 0.9135 & 0.8861 \\
                        & \tsc      & 0.9027 & 0.9119 & 0.8872 \\ \midrule
\multirow{3}{*}{\varsoft}   & \tsa      & 0.9026 & 0.9074 & 0.8878 \\
                        & \tsb      & 0.9040 & 0.9064 & 0.8830 \\
                        & \tsc      & 0.8962 & 0.9046 & 0.8808 \\ \midrule
\end{tabular}
\end{center} 
\end{table}

The \varsoft{} variant provides a ``second chance'' for other mislabeled tokens' (presumably) low quality scores (e.g.\ the second lowest) to be considered in the overall sentence quality score. For example, consider two sentences with token quality scores $[0.01, 0.99]$ and $[0.011, 0.02]$, respectively. \ourmethod\space will assign a lower score to the first sentence, but if the first token turns out not to be a label error, the sentence will become a  false positive. On the other hand, \varsoft{} also considers the second lowest token quality score, and assigns a lower score for the second sentence. In this case, it may be more likely for at least one of the tokens in the second sentence to be mislabeled vs.\ only the first token in the former sentence. Hence \varsoft{} results in attains better Lift @ \#Errors than \ourmethod{}. 
However being more dependent on all tokens' label quality scores makes \varsoft{} more sensitive to estimation error than \ourmethod{}. Thus \varsoft{} does not necessarily produce a better overall ranking of the sentences in terms of AUPRC/AUROC.

\varalt{} utilizes additional information beyond \ourmethod{}, which is estimated via Confident Learning for flagging likely label errors. Tables \ref{var:auprc} and \ref{var:auroc} show that the additions in \varalt{} and \varsoft{} can sometimes provide slight benefits to \ourmethod{}, but we do not find the performance gains to be significant enough to warrant the additional complexity (and hyperparameters) of these variants.

\clearpage
\section{Detecting Label Errors for Individual Tokens} 
\label{sec:token-level} 

Our main focus in this work was on detecting sentences that contain a mislabeled token. We focus primarily on scoring entire sentences, because many sentences anyway need to be reviewed in their entirety in order to understand an individual token is mislabeled or not, due to the need to account for the context in which this token appears.
It is still of interest to score individual tokens based on our confidence that their given label is correct, especially as this can be useful to highlight certain tokens within longer sentences. 

\begin{figure}[htp] 
    \centering
    \includegraphics[width=\textwidth]{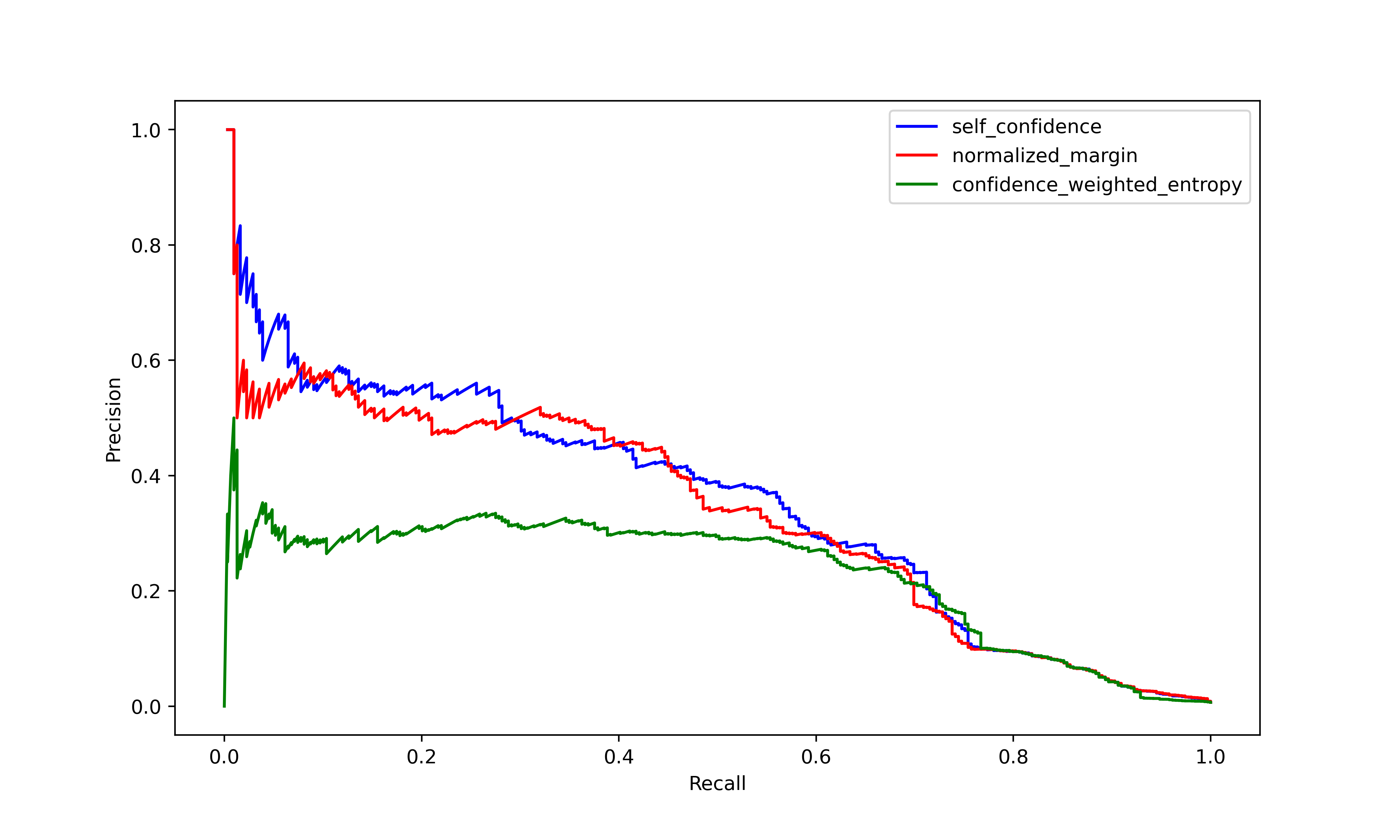}
    \caption{Precision-Recall of three label quality scoring methods applied to individual tokens rather than entire sentences. These results are from  the \texttt{bert-unmerged} setting.}
    \label{fig:token-level}
\end{figure}

Here we evaluate how well our various label quality scores are able to detect individual tokens which are mislabeled rather than which sentences contain such tokens. Each token in a sentence receives a label quality score $q_i$ based on one of the three methods described in Sec. \ref{sec:methods}: self-confidence, normalized margin, or confidence-weighted entropy. We can again evaluate these token-level $q_i$ scores against the ground truth information on whether each token's given label is actually  correct or not, this time computing precision/recall over the individual tokens independently rather than over sentences.
Figure \ref{fig:token-level} illustrates that self-confidence is generally the best option for detecting token-level label errors, as is the case for detecting sentence-level label errors with \ourmethod.

\end{document}